# Just an Update on PMING Distance for Web-based Semantic Similarity in Artificial Intelligence and Data Mining

VALENTINA FRANZONI, La Sapienza University of Rome, Italy
(franzoni@dis.uniroma1.it, valentina.franzoni@dmi.unipg.it)

**ABSTRACT**

One of the main problems that emerges in the classic approach to semantics is the difficulty in acquisition and maintenance of ontologies and semantic annotations. On the other hand, the Internet explosion and the massive diffusion of mobile smart devices lead to the creation of a worldwide system, which information is daily checked and fueled by the contribution of millions of users who interacts in a collaborative way. Search engines, continually exploring the Web, are a natural source of information on which to base a modern approach to semantic annotation. A promising idea is that it is possible to generalize the semantic similarity, under the assumption that semantically similar terms behave similarly, and define collaborative proximity measures based on the indexing information returned by search engines. The PMING Distance is a proximity measure used in data mining and information retrieval, which collaborative information express the degree of relationship between two terms, using only the number of documents returned as result for a query on a search engine. In this work, the PMINIG Distance is updated, providing a novel formal algebraic definition, which corrects previous works. The novel point of view underlines the features of the PMING to be a locally normalized linear combination of the Pointwise Mutual Information and Normalized Google Distance.
The analyzed measure dynamically reflects the collaborative change made on the web resources.

• Computing methodologies ➙ Heuristic function construction • Computing methodologies ➙ Artificial intelligence

## 1. INTRODUCTION

PMING Distance is a measure of proximity, which conveys information on relationships between two terms, e.g. word or expressions, carrying semantic meaning, used on various applications for web-based semantic similarity, such as automatic generation of significant semantic annotations for images and web objects, emotion recognition from short text [Biondi et al. 2016; Franzoni et al. 2016; Franzoni and Milani 2016; Franzoni and Milani 2014; Leung et al. 2013; Franzoni et al. 2016; Franzoni and Milani 2012; Franzoni and Milani 2015]. The automatic generation of significant semantic annotations is an important requirement for both design and implementation of web-based and information retrieval advanced services, such as query expansion, image similarity, emotion recognition, recommendation systems, et cetera, which are nowadays, more than ever, based on the management of semantic objects, rather than a simple syntactic approach.

If, on one hand, the existing semantic models are expressive enough, on the other hand they find their basic limitation on the evolution and management of ontological models and annotated content, which are not taken into account in the model itself. It results in a lack of automation capabilities and evolutionary maintenance that is highly relevant especially when you want to generate context-based semantic annotations or focus on specific social networks or repositories.

For a modern approach, search engines are the natural source of semantic information. To use the indexing content, which they provide, is a valid approach to evaluate proximity semantics of pairs of terms, or groups of terms. The general idea is to use search engines as a black box to which submit queries and extract useful statistics, to evaluate proximity semantics about the occurrence of a term or a set of terms, just counting the number of results.

This study updates the PMING, correcting previous works. The novel point of view underlines the features of the PMING to be a locally normalized linear combination of the Pointwise Mutual Information and Normalized Google Distance. Among the significant contributions of this work is the introduction of a novel formal algebraic definition of the PMING Distance, which tests carried out the best performances, obtained separately from the measures considered in literature [Franzoni and Milani 2016; Franzoni and Milani 2012].

**RELATED WORK**

PMING Distance was presented firstly in May 2012 in [Franzoni 2012], then in December 2012 in [Franzoni and Milani 2012] with a broad comparison with state-of-the-art measures and metrics, e.g. PMI, NGD, Confidence, Chi-squared in information retrieval. The aim of the work was to study the features of those measures of semantic proximity, which best adapt to the use of information provided by search engines as a base for a semantic content extraction. Among the main contributions by Franzoni, there was the systematic comparison of diverse measures, with respect to different search engines (both generalist, e.g. Bing, Google, Yahoo, and specialized, e.g. YouTube, Flickr) on a complete set of evaluation metrics, including a novel approach on clustering as a metric for evaluation of semantics, which contribute has been later deepened in a dedicated work [Franzoni and Milani 2016].

PMING Distance has been used in artificial intelligence as a heuristic in applications related to heuristic search and retrieval, e.g. heuristic semantic browsing of collaborative networks [Franzoni and Milani 2013], semantic context extraction for cooperative work in design [Franzoni and Milani 2015], multi-path traces in semantic networks [Franzoni 2017], semantic concept chaining [Franzoni and Milani 2014].

PMING Distance has also been used in data mining for automated extraction and tagging of images, in [Pallottelli et al. 2016; Leung et al. 2013], and then, recently, for semantic emotion recognition in web objects [Biondi et al. 2016; Franzoni et al.2016], where PMING is also compared with other measures for semantic extraction in the emotional domain, using the VSM for latent knowledge elicitation on the Ekman model of emotions.

**PMING DISTANCE: AN UPDATED ALGEBRAIC FORMULATION**

PMING is based on assigning a distance value on words queried in a search engine, e.g. Google, Bing, Yahoo, Flickr, YouTube, and so on, useful in any generalist or specialized search engine that indexes a high number of documents, and its effectiveness in improving methods and results for concept distance and context extraction has been proved with experiments using thousands of queries on several data sets.

We hereby give an updated formal algebraic definition of PMING.

PMING Distance of two terms x and y in a context W is defined, for $f(x) \geq f(y)$, as a function $PMING: W \times W \to [0,1]$:

$$PMING(x,y) = \rho \left[ 1 - \left( \log \frac{f(x,y)M}{f(x)f(y)} \right) \frac{1}{\mu_1} \right] + (1-\rho) \left( \frac{\log f(x) - \log f(x,y)}{(\log M - \log f(y))\mu_2} \right)$$

where:

1. $\rho$ is a parameter to balance the weight of components (e.g. $\rho = 0.3$);
2. $\mu_1$ and $\mu_2$ are constants which depend on the context of evaluation W, and can be defined as:
    - $\mu_1$=max PMI(x,y), with x,y ∈ W (PMI is the Pointwise Mutual information [Turney 2002; Franzoni and Milani 2013])
    - $\mu_2$=max normalized NGD(x,y), with x,y ∈ W (NGD is the Normalized Google Distance[Cilibrasi 2007; Franzoni 2016; Pallottelli et al. 2016])

- o Distances and proximity are interchangeable if on the same scale and normalized over [0,1], such that distance=1-proximity and proximity=1-distance

3. f(x),f(y),f(x,y) are the frequency of results of queries on search engines for the term x, the term y (occurrence) and the term (x AND y) (co-occurrence)
4. M is the total number (if known) or estimated number (if not known) of documents indexed by the search engine.

With this formulation, the author underlines how PMING conveys the features of PMI and NGD, and can be used on queries of two terms, i.e. two groups of terms, on a search engine, as a blackbox to extract the semantic meaning of the terms.

**CONCLUSIONS**

One of the main problems that emerges in the classic approach to semantics is the difficulty in acquisition and maintenance of ontologies and semantic annotations. On the other hand, the Internet explosion and the massive diffusion of mobile smart devices lead to the creation of a worldwide system. PMING Distance is a measure of proximity, which conveys information on relationships between two terms, and it conveys the features of a linear combination of PMI, semantic proximity, and NGD, semantic distance. PMING Distance has several applications in information retrieval, artificial intelligence, data mining and can be successfully used as a blackbox to obtain semantic proximity.

The formulation presented in this work underlines the algebraic features of PMING Distance as a combination of a proximity measure and a distance measure.